\documentclass[letterpaper]{article} 
\usepackage{aaai2026}  
\usepackage{times}  
\usepackage{helvet}  
\usepackage{courier}  
\usepackage[hyphens]{url}  
\usepackage{graphicx} 
\usepackage{natbib}  
\usepackage{caption} 
\usepackage{algorithm}
\usepackage{algorithmic}

\usepackage{newfloat}
\usepackage{listings}
\DeclareCaptionStyle{ruled}{labelfont=normalfont,labelsep=colon,strut=off}
\floatstyle{ruled}
\newfloat{listing}{tb}{lst}{}
\floatname{listing}{Listing}

\title{Schema on the Inside: A Two-Phase Fine-Tuning Method for High-Efficiency Text-to-SQL at Scale}

\author{
    Chinmay Soni\equalcontrib,
    Shivam Chourasia\equalcontrib,
    Gaurav Kumar\equalcontrib,
    Hitesh Kapoor
}

\affiliations{
    Sporta Technologies Private Limited, Mumbai, Maharashtra, India\\
    \{chinmay.soni@dream11.com, shivam.chourasia@dream11.com, gaurav.kumar@dream11.com, hitesh.kapoor@dream11.com\}
}

\begin{document}

\maketitle

\begin{abstract}
Applying large, proprietary API-based language models to text-to-SQL tasks poses a significant industry challenge: reliance on massive, schema-heavy prompts results in prohibitive per-token API costs and high latency, hindering scalable production deployment. We present a specialized, self-hosted \textbf{8B-parameter model} designed for a conversational bot in CriQ, a sister app to Dream11---India’s largest fantasy sports platform with over 250 million users---that answers user queries about cricket statistics. Our novel two-phase supervised fine-tuning approach enables the model to internalize the entire database schema, eliminating the need for long-context prompts. This reduces input tokens by over 99\%, from a 17k-token baseline to fewer than 100, and replaces costly external API calls with efficient local inference. The resulting system achieves \textbf{98.4\% execution success} and \textbf{92.5\% semantic accuracy}, substantially outperforming a prompt-engineered baseline using \textbf{Google's Gemini Flash 2.0} (95.6\% execution, 89.4\% semantic accuracy). These results demonstrate a practical path toward high-precision, low-latency text-to-SQL applications using domain-specialized, self-hosted language models in large-scale production environments.
\end{abstract}

\section{Introduction}
\begin{figure}[htbp]
\centering
\includegraphics[width=0.98\columnwidth]{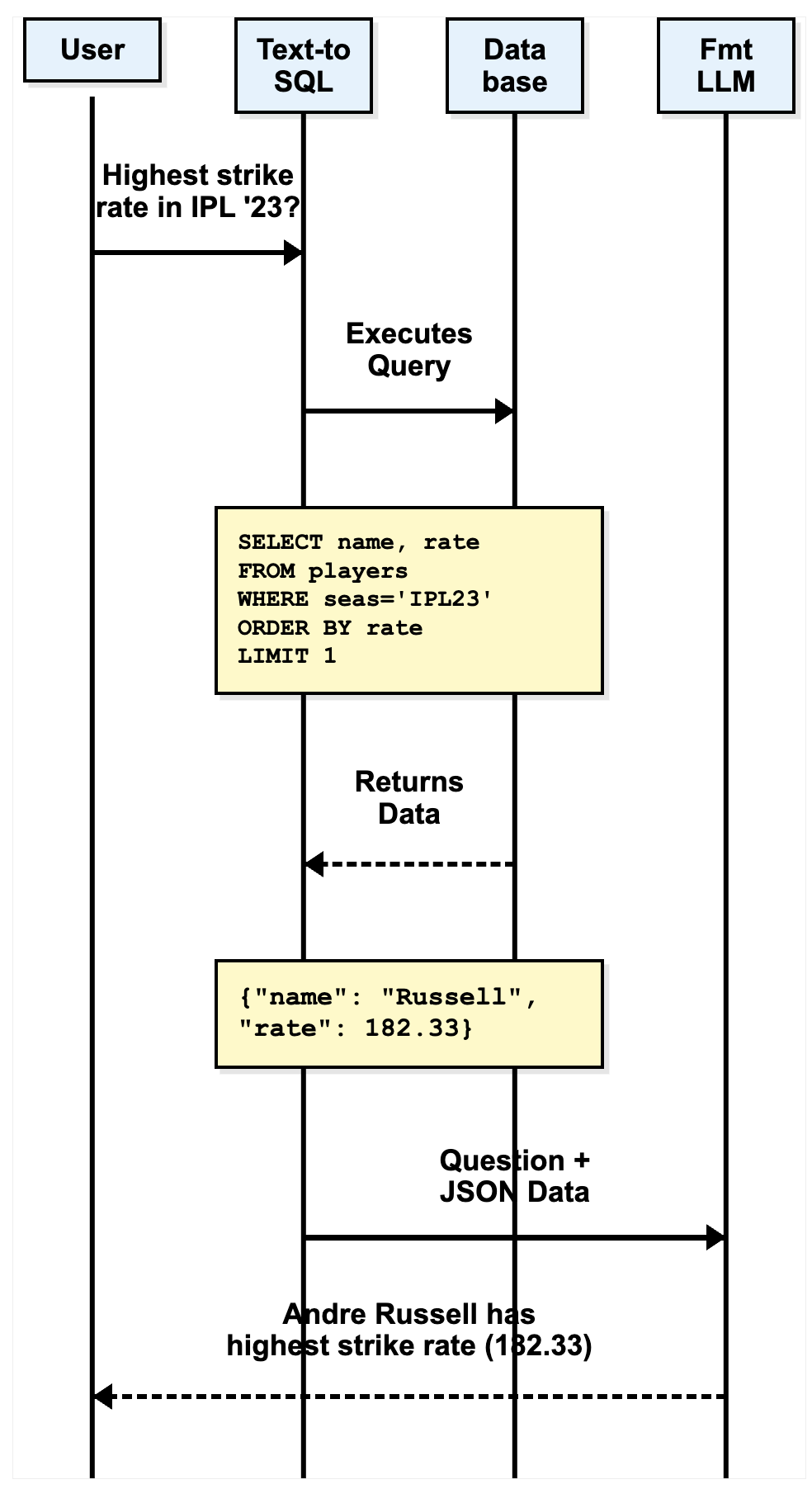}
\caption{Conceptual overview of the conversational AI feature. A user asks a question in natural language, and the system returns a data-driven answer by generating and executing a SQL query.}
\label{fig:concept}
\end{figure}

The promise of democratizing data access, moving beyond expert-driven analytics to intuitive self-service interfaces, is a key goal for modern data-intensive applications. As user expectations shift towards seamless, conversational interactions, the pressure to deliver performant and intelligent data access tools has intensified. Our team confronted these challenges directly while developing a new feature for our sports intelligence application: a conversational AI assistant that could effortlessly query our complex cricket database, answering any statistical question in plain English (Figure \ref{fig:concept}).

\begin{figure}[htbp]
\centering
\includegraphics[height=8cm,width=0.82\columnwidth]{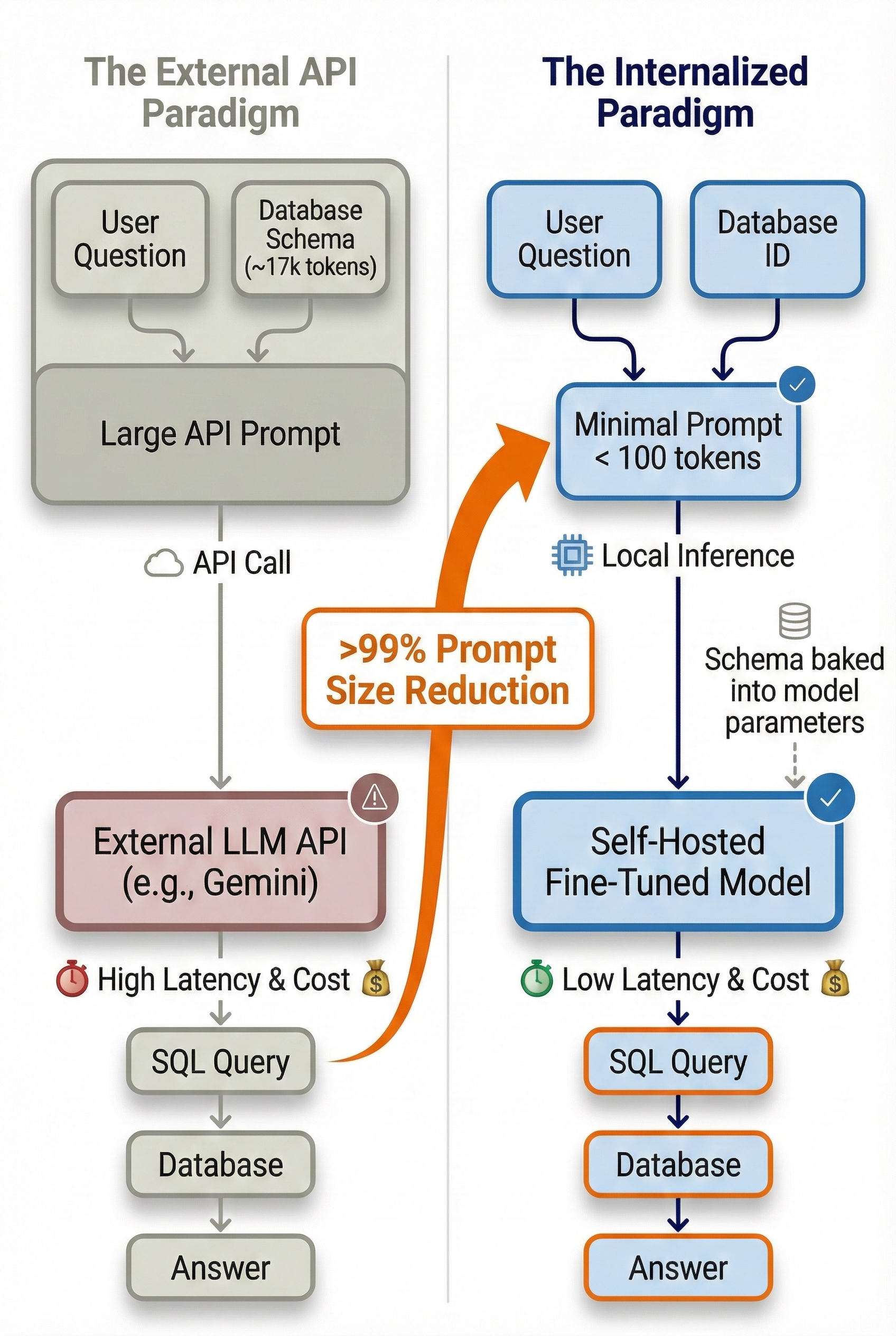}
\caption{A visual comparison of the two paradigms : \\
\textbf{Left side :} ``The External API Paradigm", where a large prompt is sent to a third-party, pay-per-token service. \\
\textbf{Right side :} ``The Internalized Paradigm", showing a small prompt with just the question going to a fine-tuned, self-hosted model with the schema ``baked in". 
}

\label{fig:paradigm_comparison}
\end{figure}

A dominant industry paradigm for enabling this is using powerful, general-purpose Large Language Models (LLMs) via third-party APIs. This approach typically involves feeding the external model a massive prompt containing the user's question along with the full database schema and a handful of examples (Figure \ref{fig:paradigm_comparison}, left). However, this \textbf{prompt-heavy approach} creates significant, often, insurmountable barriers to production deployment. Stuffing the context window with thousands of tokens for every user query is computationally wasteful, leading to prohibitive operational costs (as most API providers charge per token) and high user-facing latency \citep{hooper2024squeezed}. This architectural constraint resulted in an unacceptably sluggish user experience, with query latency frequently exceeding several seconds, leading to increased query abandonment. Furthermore, the initial prototype suffered from an inconsistent success rate of only 82\% (execution accuracy of generated SQL). These issues highlighted the need for a more cost-effective and reliable solution.

Our initial prototype, leveraging a third-party LLM API, specifically Gemini Flash \cite{google:23:gemini}, starkly illustrated the profound limitations of off-the-shelf solutions for our critical requirements. Despite Gemini Flash's well-earned reputation as a leader in low-latency inference, the complex demands of our queries, often requiring a substantial information and cricket context encoded in prompt to encapsulate the necessary context and schema, inherently bottlenecked performance. These critical drawbacks necessitated a fundamental re-architecture. We decisively pivoted from this generic, third-party API approach to develop a specialized, in-house asset: a highly-optimized, fine-tuned model trained using a novel two-phase methodology (Figure \ref{fig:paradigm_comparison}, right).

This work addresses the critical trade-off between accuracy, cost, and latency. We build upon the concept of schema internalization via fine-tuning, an idea explored in works like YORO \citep{sun:24, kobayashi:24:yoro}, which demonstrated a model’s ability to learn a database structure. Our primary contribution is a novel and practical \textbf{two-phase fine-tuning recipe} specifically designed for production environments. While prior work established the possibility of internalization, our method provides a concrete, structured curriculum that proved essential for enabling a model to reliably recall and apply schema knowledge from a minimal prompt. This approach creates a smaller, low-latency model that beats the accuracy of a large-context system without the associated costs, making it ideal for user-facing products.

Our contributions are:
\begin{itemize}
\item \textbf{A Novel Two-Phase Fine-Tuning Technique for Schema Internalization:} We introduce and validate a new supervised fine-tuning recipe for text-to-SQL that achieves 98.4\% execution accuracy while reducing prompt tokens by over 99\%. Our method, which explicitly teaches a model to both learn the SQL generation task in full context and to recall the schema from memory, is our core technical contribution.
\item \textbf{A Comprehensive Case Study on Productionizing Text-to-SQL:} We outline the engineering journey of enhancing our Statsbot, from a limited prompt-based prototype to a deployed, fine-tuned model. This provides a blueprint for balancing accuracy, latency, and cost in user-facing text-to-SQL applications.
\end{itemize}

\section{Related Work}

\paragraph{Early and Deep Learning Approaches.}
Initial text-to-SQL research relied on rule-based and statistical methods, as exemplified by early work like \citep{dong:16:lang2logic}. The advent of deep learning brought sequence-to-sequence (seq2seq) models, with Seq2SQL \citep{zhong:17:seq2sql} being a notable early example. Addressing challenges such as the "lexical gap" (where natural language queries use different terminology than the database schema), subsequent approaches introduced innovations like SQLNet's slot-filling mechanism \citep{xu:17:sqlnet} and the use of intermediate representations such as NatSQL \citep{gan:21:natsql}. A significant breakthrough was RAT-SQL \citep{wang:20}, which introduced relation-aware self-attention, greatly improving performance by explicitly modeling the relationships between query tokens and database schema elements. Further advancements included semi-autoregressive parsing \citep{rubin:20:smbop}, grammar-augmented pre-training \citep{yu:21:grappa}, and methods that decoupled the text-to-SQL task into sub-problems \citep{li:23:resdsql, li:23:graphix}. This progress was significantly accelerated by the development of challenging benchmark datasets, including WikiSQL \citep{zhong:17:wikisql}, Spider \citep{yu:18}, KaggleDBQA \citep{lee:21:kaggledbqa}, BIRD \citep{li:23:bird}, and Dr.Spider \citep{chang:23:drspider}.

\paragraph{The Rise of Large Language Models (LLMs).}
The introduction of transformer-based LLMs, such as T5 \citep{raffel:20} and BART \citep{lewis:20:bart}, fundamentally reframed text-to-SQL as a text-to-text generation problem. This paradigm shift was further propelled by models like GPT-3 \citep{brown:20:gpt3}, which popularized in-context learning (ICL). ICL-based systems harness the powerful reasoning capabilities of LLMs through elaborate prompts. Techniques like Chain-of-Thought (CoT) prompting \citep{wei:22:cot, tai:23:cot} and self-correction mechanisms \citep{chen:23:selfdebug}, exemplified by DIN-SQL \citep{pourreza:23}, have achieved state-of-the-art results in text-to-SQL. However, these ICL-heavy approaches, which often necessitate extensive prompt engineering, come with significant drawbacks, including prohibitive latency and cost, making them impractical for real-time industrial applications. For instance, DIN-SQL with GPT-4 has reported a 60-second latency and a cost of \$0.50 per query \citep{pourreza:23}. This limitation has spurred the development of capable open-source models like LLaMA \citep{touvron:23:llama1, touvron:23:llama2}, specialized code generation models such as StarCoder \citep{li:23:starcoder} and CodeLlama \citep{roziere:23:codellama}, and SQL-centric pre-trained models like the CODES series \citep{li:24:codes}.

\paragraph{Schema Internalization and PEFT.}
Our work directly addresses the performance bottleneck inherent in ICL approaches by introducing a novel strategy: \textbf{schema internalization}. This technique encodes the database schema directly into the model's parameters, enabling accurate recall with only a minimal trigger in the prompt. This concept is inspired by the YORO paradigm \citep{sun:24, kobayashi:24:yoro}, which advocates for training ``expert models" through fine-tuning on database-specific question-SQL pairs, often generated via self-instruction \citep{wang:23:selfinstruct}. This process effectively ``bakes in" the schema knowledge, thereby eliminating the need to explicitly provide the schema in the prompt during inference. While the YORO paradigm inspired our work by demonstrating the potential of schema internalization to reduce inference costs and input length, our approach provides a distinct and practical recipe for production environments. Specifically, we differentiate ourselves by not requiring skeleton-based synthetic data, which simplifies the data preparation process.

\section{Application Description: The Sports Intelligence Platform}
Our work is centered on \textbf{CriQ}, a comprehensive sports intelligence application from Sporta Technologies Private Limited, the parent company of \textbf{Dream11}—India's largest fantasy sports platform with over 250 million registered users. While CriQ itself currently serves a rapidly growing base of 350,000 logged-in users, it is architected to handle the immense scale of the broader Dream11 ecosystem. This context underscores the critical need for a highly scalable and cost-effective text-to-SQL solution.

A core component of the CriQ platform is its conversational AI assistant—that answers user queries about cricket statistics, the optimization of which is the focus of this paper. The goal of this feature is to provide a simple, conversational interface to our complex cricket database. This database covers everything from player-specific career statistics and match-by-match performance logs to detailed ball-by-ball event data and venue characteristics. The tables are interconnected through a web of foreign key relationships, making manual query construction a task for expert analysts only.

The conversational assistant must handle this complexity seamlessly. During peak periods, such as major tournaments, the feature must be capable of handling significant concurrent requests. Users ask a wide range of questions, from simple lookups like, ``\textit{Who has the highest strike rate in IPL 2023?}" to complex analytical queries like, ``\textit{Compare Rohit Sharma's batting average in run-chases when setting a target}." The system's \textbf{query generation engine} must convert this natural language input into a precise, executable PostgreSQL query, run it against our production database, and return a human-readable answer. The database schema, while comprehensive, is static---a key characteristic that makes our schema internalization approach viable and effective. The high user engagement with this feature underscores the importance of its performance, accuracy, and cost.

\section{Our Fine-Tuning Methodology}
The core of the improved conversational AI feature is its new fine-tuned model, developed through a deliberate, multi-component process. This section details the steps taken to prepare the data, establish baselines, and implement our novel two-phase fine-tuning approach.

\subsection{Data Preparation and Baseline Establishment}
\begin{figure}[t!]
\centering
\includegraphics[width=1\columnwidth]{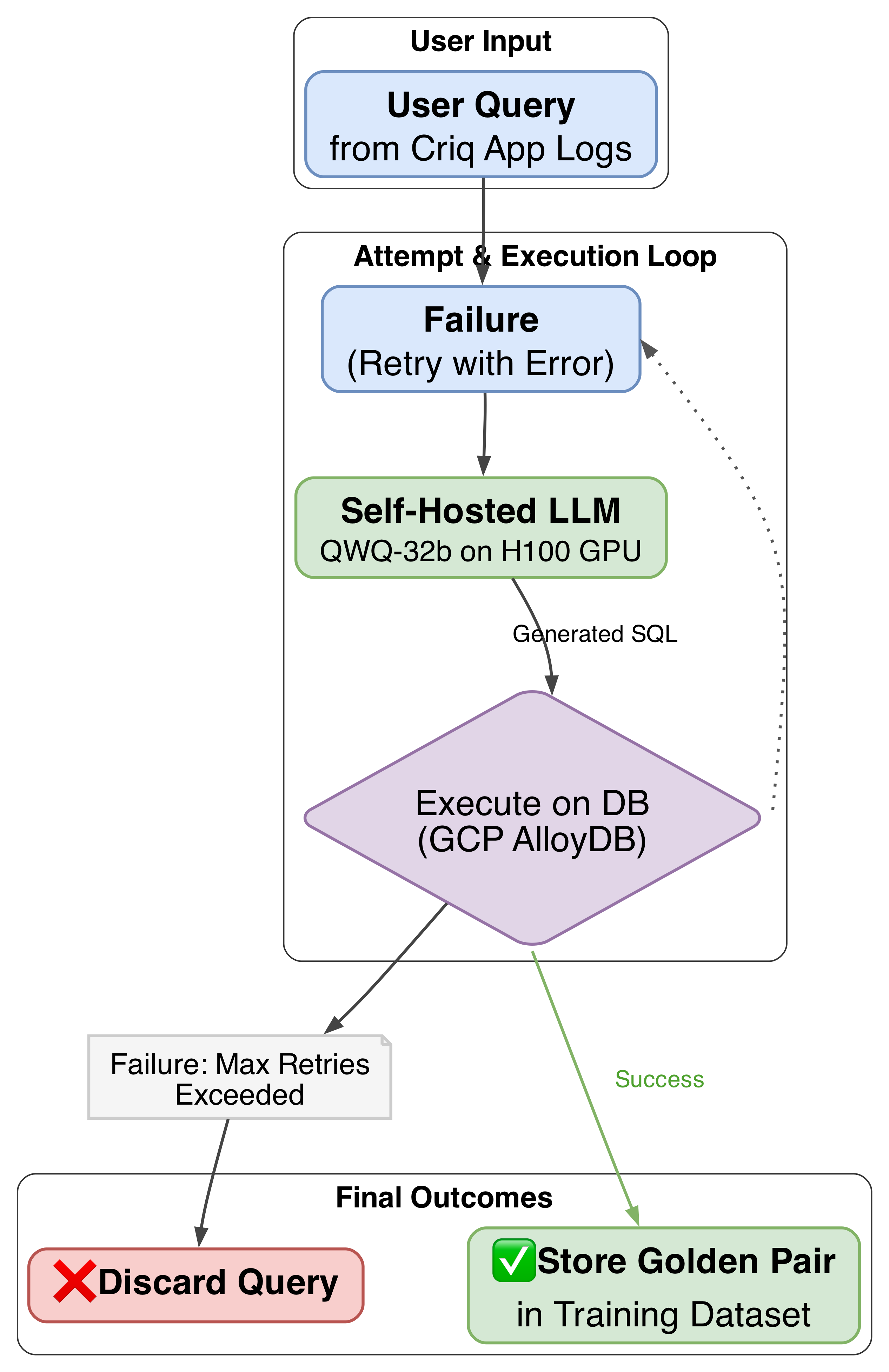}
\caption{The iterative process for generating the ground-truth dataset. User queries are processed by a self-hosted reasoning LLM to generate SQL. The query is executed, and if successful, the question-SQL pair is stored. If it fails, the prompt is updated with the error, and the process is retried up to a maximum number of attempts.}
\label{fig:data_gen_loop}
\end{figure}

\textbf{Ground-Truth Data Generation:} We began by sampling 11,000 unique user queries from our application logs. To create a high-quality, ground-truth dataset for these queries, we employed a powerful, self-hosted open-source reasoning model, \textbf{Qwen/QwQ 32B} \citep{bai:23:qwen}. While highly effective for complex reasoning tasks and data generation, its inherent high latency made it unsuitable for our real-time, user-facing production environment. This model was used in an iterative offline loop to regenerate correct question-SQL pairs (Figure \ref{fig:data_gen_loop}). If the generated SQL failed execution, the prompt was amended with the database error, and the model would attempt to self-correct. This process was highly effective for creating a high-quality training set and served as an infinite tap for synthetic data generation.

\paragraph{Initial Model Benchmarking.} To select a strong foundation for our work, we fine-tuned several open-source models in the ~8B parameter range on our ground-truth dataset. Their performance was evaluated on a holdout set of user queries based on Execution Accuracy, which measures the percentage of generated SQL queries that execute successfully against the database without raising an error. The results, presented in Table \ref{tab:initial_benchmark}, indicated that the Qwen model family demonstrated superior performance. Consequently, the Qwen 3 8B model, which achieved the highest accuracy (92.61\%), was selected for all subsequent experiments.

\begin{table}[t!]
\centering
\begin{tabular}{lc}
\hline
\textbf{Model} & \textbf{Exec. Acc.} \\
\hline
LLaMA 3 8B \citep{meta:24:llama3} & 83.76\% \\
CodeLLAMA 7B \citep{roziere:23:codellama} & 84.98\% \\
Qwen 2.5 7B \citep{bai:23:qwen} & 88.61\% \\
Qwen 3 8B \citep{bai:23:qwen} & 92.61\% \\
\hline
\end{tabular}
\caption{Initial fine-tuning benchmark results. Based on its superior performance, the Qwen 3 8B model was selected for all subsequent iterations.}
\label{tab:initial_benchmark}
\end{table}

\subsection{Prompt Engineering and Baseline Definition}
A deep audit of thousands of failed queries from our initial benchmarks revealed recurring error patterns (e.g., incorrect table joins, misuse of aggregate functions, breakages in nested subqueries). We codified a set of best practices to correct these. For instance, to handle ambiguous filters that could apply to multiple joined tables, we enforced the use of Common Table Expressions (CTEs) to pre-filter data, making the final query logic cleaner and less error-prone.

These failure patterns and correction principles were fed into Gemini 2.5 Pro, which generated an improved, more robust system prompt. The result was a highly-engineered ~17k-token prompt that included not just the schema, but also execution-oriented templates and explicit failure-avoidance instructions. When used with the original Gemini 2.0 Flash model, this new prompt boosted execution accuracy to 95.6\%. This process established two distinct baselines for comparison: the initial prototype and a new, much stronger (but costly) prompt-engineered baseline (Table \ref{tab:baselines}).

\begin{table}[t!]
\centering
\begin{tabular}{lcc}
\hline
\textbf{System} & \textbf{Prompt Tokens} & \textbf{Exec. Acc.} \\
\hline
Initial Prototype & ~10k & 82.0\% \\
Prompt-Eng. Baseline & ~17k & 95.6\% \\
\hline
\end{tabular}
\caption{Comparison of the initial prototype and the improved, prompt-engineered baseline. Both used the Gemini 2.0 Flash model via its third-party API.}
\label{tab:baselines}
\end{table}

\subsection{The Two-Phase Fine-Tuning Recipe for Schema Internalization}
The central challenge in moving away from the prompt-heavy paradigm is enabling a model to generate accurate, schema-aware SQL queries without being given the full schema in every prompt. Our solution is to make the model internalize the schema so it becomes part of its parametric knowledge. To achieve this, we developed a novel, two-phase fine-tuning regime designed to first teach the schema and then train the model to recall and apply that knowledge on command.

We build on the schema internalization paradigm explored by works like YORO \cite{kobayashi:24:yoro}, which demonstrated that a model can be trained to associate a full database's structure with a simple identifier. We adopt this core concept by assigning a unique ID to our database, \textbf{cricket\_stats\_db\_v1}. This ID acts as a recall trigger: instead of providing thousands of tokens of schema information, the final prompt only needs to contain this short string. The model, having been specially trained, recognizes this trigger and accesses the complete, memorized schema from its internal weights to correctly process the user's query. Our primary contribution is the specific two-phase curriculum that makes this internalization process robust and highly accurate for a real-world database.

\begin{figure*}[t!]
\centering
\includegraphics[width=\textwidth]{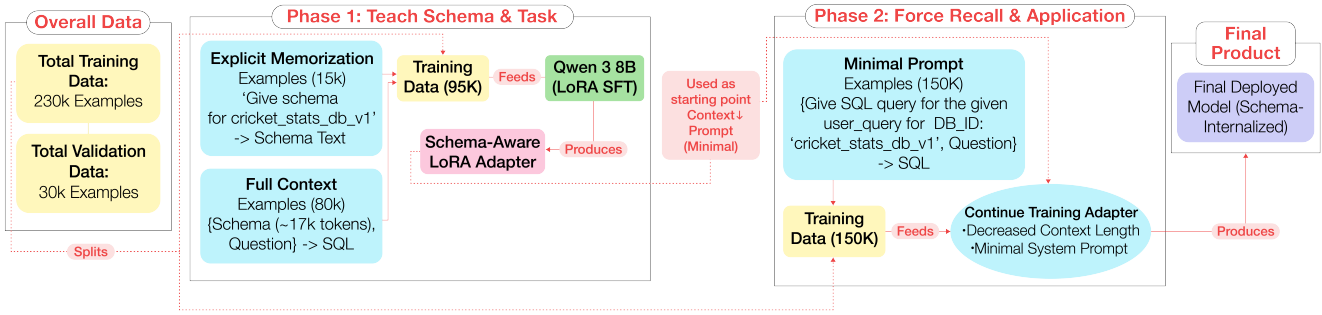}
\caption{An overview of the two-phase fine-tuning methodology. Phase 1 combines full-context examples with explicit schema memorization tasks to create a schema-aware LoRA adapter. Phase 2 continues training using this adapter but with a minimal prompt, forcing the model to rely on its internalized knowledge of the schema.}
\label{fig:two_phase_training}
\end{figure*}

\paragraph{Phase 1: Schema Learning and Association.}
This phase acts as a comprehensive ``schooling" for the model, fine-tuning it on a mixed dataset combining two distinct task types essential for success:
\begin{itemize}
\item \textbf{SQL Generation from Full Context (80,000 examples):} In this primary task, the model learns the complex reasoning of translating natural language to SQL. By training on the full 17k-token prompt, which included both the database schema and cricket-specific context, it learns how to correctly join tables, apply filters, and structure queries.
\item \textbf{Explicit Schema Memorization (15,000 examples):} In this secondary task, the model learns to simply recall the schema itself. The input was a prompt like, ``Give me full schema details for database\_id = `cricket\_stats\_db\_v1'", and the expected output was the full text of the database schema. This task explicitly forces the model to encode the schema's structure and vocabulary into its parameters, associating it with the unique database ID.
\end{itemize}

\paragraph{Phase 2: Schema Recall and Application.}
The objective of this phase is to make the model use its internalized knowledge. By removing the ``crutch" of the full-context prompt, the model is forced to rely on the knowledge it encoded during Phase 1. To achieve this, the LoRA adapter from Phase 1 was further trained on 150,000 query examples using only a minimal prompt, which contained nothing but the user's question and the database ID trigger.

\begin{quote}
\textbf{Example Minimal Prompt for Phase 2:}\\
User Query: ``Highest batting average player in IPL?"\\
Generate SQL for user query using database\_id: ``cricket\_stats\_db\_v1"
\end{quote}

\section{Training and Implementation Details}
All fine-tuning experiments were conducted using LoRA (Low-Rank Adaptation) \citep{hu:21}, a cornerstone of Parameter-Efficient Fine-Tuning (PEFT). We chose PEFT because it dramatically reduces the memory footprint and enables faster iteration. Instead of training the model's 8 billion original parameters, LoRA freezes the pre-trained weights and injects much smaller, trainable low-rank matrices into the attention layers of the network. For our experiments, we used a LoRA rank (`r') of 64 and a scaling factor (`alpha') of 128, targeting the query, key, and value projection layers (`q\_proj', `k\_proj', `v\_proj') of the model.

Our training runs were conducted on a cluster of 8x NVIDIA H100 80GB GPUs. We utilized Distributed Data Parallel (DDP) to accelerate training across the GPUs. For the optimizer, we used AdamW with a learning rate of 2e-5 and a cosine learning rate scheduler. A key optimization in our workflow was resuming training on the same LoRA adapters between phases. After completing the high-context training in Phase 1 (with a context length of 20,480 tokens), we began Phase 2 by loading the resulting adapters and simply reducing the context length to 1,024 for the minimal-prompt examples. This seamless transition significantly sped up experimentation.

For deployment, the final fine-tuned model is self-hosted on our own GPU-enabled machines and served using vLLM \cite{kwon2023vllm}, an open-source library designed for fast and memory-efficient LLM inference. vLLM's use of Paged-Attention was particularly beneficial, allowing for higher batch sizes and significantly increasing our requests-per-minute throughput compared to standard Hugging Face implementations.

\section{Deployment and Impact Analysis}
The new fine-tuned model was deployed into production, replacing the prompt-engineered Gemini system for the conversational AI feature in our application.

\subsection{Evaluation Setup}
\begin{itemize}
\item \textbf{Model:} Our final model based on the Qwen architecture, fine-tuned with the two-phase recipe.
\item \textbf{Baseline:} The prompt-engineered baseline model (Gemini 2.0 Flash with the ~17k-token prompt, accessed via its API).
\item \textbf{Test Set:} A holdout set of 30k curated, real-world user queries, unseen during training.
\item \textbf{Deployment Infrastructure:} The final model is self-hosted on our own GPU-enabled machines and served using vLLM.
\item \textbf{Metrics:}
\begin{itemize}
\item \textbf{Execution Success:} The percentage of generated SQL queries that execute without any database errors. This is a measure of syntactic and structural correctness.
\item \textbf{Semantic Accuracy:} The percentage of successfully executed queries that return the correct data to answer the user's question. This is a measure of correctness, evaluated using an LLM via batch api.
\item \textbf{Total Prompt Length:} The total number of tokens in the input prompt sent to the model.
\end{itemize}
\end{itemize}

\subsection{Impact on Accuracy and Efficiency}
\begin{figure}[htbp]
\centering
\includegraphics[width=1\columnwidth]{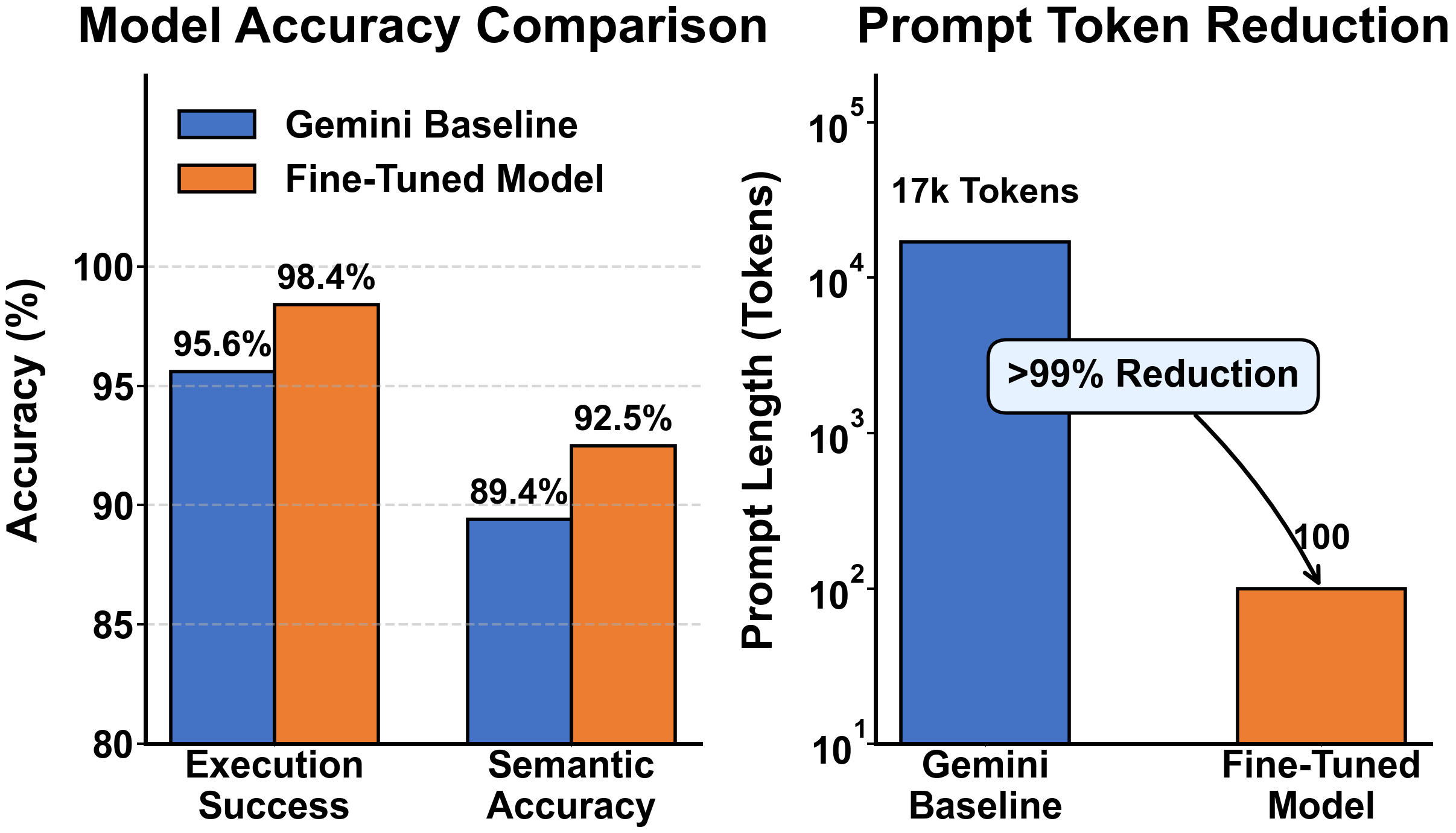}
\caption{Performance comparison of the final deployed model against the prompt-engineered Gemini baseline. The left chart shows our model achieved higher execution success (98.4\% vs 95.6\%) and semantic accuracy (92.5\% vs 89.4\%). The right chart illustrates the dramatic ($>99\%$) reduction in prompt token length.}
\label{fig:final_results}
\end{figure}

The deployment of the new model resulted in significant, measurable improvements across all key business and performance metrics, as detailed in Figure \ref{fig:final_results}. The final model is not only more accurate but also dramatically more efficient.

Economically, the impact was even more pronounced. The shift from a variable, per-query API cost to a fixed infrastructure cost.

\section{Ablation Studies}
To rigorously validate our final two-phase methodology, we conducted ablation studies to justify our core design choices.

\subsection{Study 1: Necessity of the Two-Phase Curriculum}
To demonstrate the value of our two-phase curriculum, we compared its performance against a simpler, single-phase fine-tuning approach. In this simpler approach, the model was trained from the beginning using only minimal prompts (the user question and database ID), without the benefit of seeing the full-context examples in Phase 1.

The results were stark: the single-phase model achieved only \textbf{74.96\% execution accuracy}. This is significantly lower than the \textbf{98.4\%} achieved by our two-phase recipe. This confirms our hypothesis that the initial phase of training with full-context prompts is essential. It acts as a necessary ``teaching" phase for the model to learn the complex text-to-SQL task space before it can be expected to perform well from a minimal prompt.

\subsection{Study 2: The Importance of Explicit Schema Memorization}
The impact of explicit schema memorization in Phase 1 is significant. To quantify its effect, we trained the model using the exact same two-phase process but excluded the 15,000 memorization examples, relying solely on the 80,000 full-context SQL examples.

Without the explicit memorization task, the model's performance dropped to 87.2\% execution accuracy and 79.5\% semantic accuracy. This result is critical: it demonstrates that while the model can learn the schema implicitly from full-context SQL examples, this implicit knowledge is not robust enough for reliable recall. The explicit memorization task is crucial for solidifying the model’s internal representation of the schema, enabling it to achieve the final ($>98\%$) execution accuracy when prompted with only the trigger ID in Phase 2.

\subsection{Study 3: Impact of Model Scale}
To determine the best model size for our application, we ran our full two‑phase fine‑tuning pipeline on three Qwen models with different parameter counts.

\begin{figure}[htbp]
\centering
\includegraphics[width=1\columnwidth]{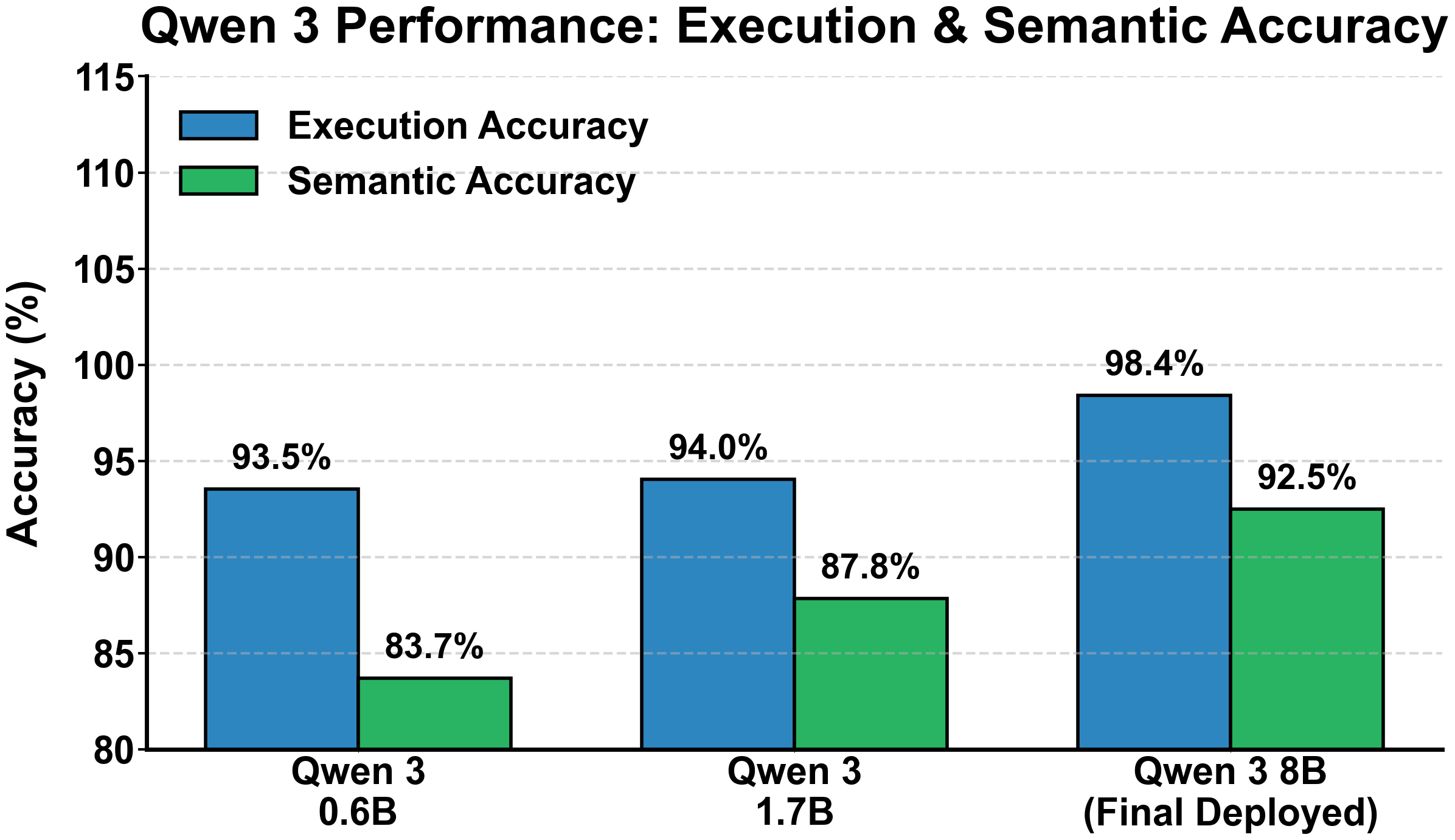}
\caption{Execution and semantic accuracy for different Qwen 3 model sizes fine-tuned with our two-phase methodology. The 8B parameter model shows the highest accuracy, justifying its selection.}
\label{fig:ablation_scale}
\end{figure}

Figure \ref{fig:ablation_scale} shows the expected pattern: accuracy rises with model size. While smaller models reduce latency and hosting costs, their performance levels off below the accuracy threshold our user‑facing feature demands. Because semantic precision is essential for user trust and overall reliability, we cannot compromise on accuracy.

The 8‑billion‑parameter variant offers the optimal trade‑off. It delivers the highest accuracy of the group while keeping inference latency and compute requirements within a cost‑effective range for large‑scale self‑hosting. In practice, models with fewer than roughly 8B parameters can internalize some schema information, but only an 8B‑class model achieves the state‑of‑the‑art accuracy we target.

\section{Discussion and Industrial Impact}
Our journey underscores the value of a methodical, iterative approach to building practical text-to-SQL solution. By establishing strong baselines, performing deep failure analysis, and finally developing a targeted fine-tuning strategy validated through rigorous ablation, we systematically addressed the dual challenges of accuracy and efficiency.

The industrial impact is twofold:
\begin{itemize}
\item \textbf{Shift in Cost Structure:} The primary benefit is a shift from a high, variable per-query third-party API cost to a predictable, fixed infrastructure cost. By self-hosting an optimized open-source model, we can handle high traffic volumes without the runaway expense of calling a proprietary API with a 17k-token prompt for every request.
\item \textbf{Enhanced Reliability:} The increase in semantic accuracy means users receive the correct information more often, building trust in the application and its AI-powered features. This is paramount for any analytical tool.
\end{itemize}
This process confirms that for specialized, high-throughput applications with fixed schema, investing in fine-tuning to internalize domain knowledge is a superior long-term strategy to relying on ever-larger prompts. While the initial investment in data generation and fine-tuning is non-trivial, the long-term benefits in cost, performance, and user experience provide a clear return. Our approach is generalizable to other domains where a large but static context (e.g., API documentation, policy manuals, game rule-books) must be consistently applied. A key limitation, however, is the static nature of the schema; significant changes to the database would require a new fine-tuning cycle.

\section{Conclusion}
We have presented a case study on the successful, multi-stage optimization of a key conversational feature within our sports intelligence application. By progressing from an initial prototype to a prompt-engineered system and finally to a highly optimized, schema-internalized model, we demonstrated a clear path to building text-to-SQL based applications that are accurate, fast, and cost-effective. Our novel two-phase fine-tuning strategy, validated through rigorous ablation studies, enabled the final model to surpass a strong, prompt-based baseline on all key metrics. This methodology provides a practical and effective blueprint for other engineering teams looking to move beyond prototype-level LLM implementations to build robust, production-ready AI features that can operate at scale without prohibitive costs.

\section{Future Work}
While our model is highly accurate, the remaining failures are often due to subtle issues like ambiguous column aliasing or complex temporal reasoning. A promising direction for future work is to use preference-based fine-tuning methods like Direct Preference Optimization (DPO) \citep{rafailov:23:dpo} to refine the model on these nuanced cases. By training the model to prefer correct, well-structured queries over subtly incorrect ones, we could further push the boundaries of accuracy and reliability for the conversational AI feature.

\bibliography{aaai2026}

\end{document}